\pdfoutput=1

\documentclass[journal,twoside,web]{ieeecolor}

\usepackage{tmi}
\usepackage{times}
\usepackage{epsfig}
\usepackage{graphicx}
\usepackage{amsmath}
\usepackage{amssymb}
\usepackage{authblk}
\usepackage[utf8]{inputenc}
\usepackage{placeins}
\usepackage{tabu}
\usepackage{booktabs}
\usepackage{subcaption}
\usepackage[pagebackref=true,breaklinks=true,letterpaper=true,colorlinks,bookmarks=false]{hyperref}

\def\BibTeX{{\rm B\kern-.05em{\sc i\kern-.025em b}\kern-.08em
    T\kern-.1667em\lower.7ex\hbox{E}\kern-.125emX}}
\begin{document}
\bstctlcite{IEEEexample:BSTcontrol}

\title{Towards annotation-efficient segmentation via image-to-image translation}
\author{Eugene Vorontsov, Pavlo Molchanov, Christopher Beckham, 
%Francisco Romero, Wonmin Byeon, Shalini De Mello, Ming-Yu Liu,
Jan Kautz, and Samuel Kadoury

% \thanks{Submitted August 18, 2020. This work was supported by NVIDIA Corporation, a Natural Sciences and Engineering Research Council Doctoral Postgraduate Scholarship (NSERC PGS-D, Eugene Vorontsov), and Oncotech grand 293740, involving Fonds de la recherche en santé du Québec (FRQS), MEDTEQ, and Elekta.}
% \thanks{E. Vorontsov is with Ecole Polytechnique de Montreal, Montreal, Quebec H3T1J4 Canada and Mila (email: eugene.vorontsov@gmail.com). The bulk of this work was done while E. Vorontsov was interning at NVIDIA.}
% \thanks{P. Molchanov is with NVIDIA Corporation in Santa Clara, California 95051 USA (email: pmolchanov@nvidia.com).}
% \thanks{C. Beckham is with Ecole Polytechnique de Montreal, Montreal, Quebec H3T1J4 Canada and Mila (email: christopher.j.beckham@gmail.com).}
% \thanks{F. Romero is with Ecole Polytechnique de Montreal, Montreal, Quebec, H3T1J4 Canada (email: francisco.perdigon@polymtl.ca).}
% \thanks{W. Byeon is with NVIDIA Corporation in Santa Clara, California 95051 USA (email: wbyeon@nvidia.com).}
% \thanks{S. De Mello is with NVIDIA Corporation in Santa Clara, California 95051 USA (email: shalinig@nvidia.com)}
% \thanks{M. Liu is with NVIDIA Corporation in Santa Clara, California 95051 USA (email: mingyul@nvidia.com)}
% \thanks{J. Kautz is with NVIDIA Corporation in Santa Clara, California 95051 USA (email: jkautz@nvidia.com)}
% \thanks{S. Kadoury is with with Ecole Polytechnique de Montreal, Montreal, Quebec H3T1J4 Canada and Centre hospitalier de l'Université de Montréal (CHUM; email: samuel.kadoury@polymtl.ca).}

\thanks{Submitted December 21, 2020. This work was supported by NVIDIA Corporation, a Natural Sciences and Engineering Research Council Doctoral Postgraduate Scholarship (NSERC PGS-D, Eugene Vorontsov), and Oncotech grant 293740, involving Fonds de la recherche en santé du Québec (FRQS), MEDTEQ, and Elekta.}
\thanks{E. Vorontsov is with Ecole Polytechnique de Montreal and Mila (eugene.vorontsov@gmail.com). The majority of this work was done while E. Vorontsov was interning at NVIDIA.}
\thanks{P. Molchanov is with NVIDIA Corporation (pmolchanov@nvidia.com).}
\thanks{C. Beckham is with Ecole Polytechnique de Montreal and Mila ( christopher.j.beckham@gmail.com).}
% \thanks{F. Romero is with Ecole Polytechnique de Montreal (email: francisco.perdigon@polymtl.ca).}
% \thanks{W. Byeon is with NVIDIA Corporation (email: wbyeon@nvidia.com).}
% \thanks{S. De Mello is with NVIDIA Corporation (email: shalinig@nvidia.com)}
% \thanks{M. Liu is with NVIDIA Corporation (email: mingyul@nvidia.com)}
\thanks{J. Kautz is with NVIDIA Corporation (jkautz@nvidia.com)}
\thanks{S. Kadoury is with with Polytechnique Montreal and Centre hospitalier de l'Université de Montréal (samuel.kadoury@polymtl.ca).}
\thanks{Polytechnique  Montreal is at Montreal, Quebec H3T1J4 Canada.}
\thanks{NVIDIA Corporation is at Santa Clara, California 95051 USA.}
}

\maketitle

%%%%%%%%% ABSTRACT
\begin{abstract}
Often in medical imaging, it is prohibitively challenging to produce enough boundary annotations to train deep neural networks for accurate tumor segmentation. We propose the use of weak labels about whether an image presents tumor or whether it is absent to extend training over images that lack these annotations. Specifically, we propose a semi-supervised framework that employs unpaired image-to-image translation between two domains, presence vs.\ absence of cancer, as the unsupervised objective. We conjecture that translation helps segmentation---both require the target to be separated from the background. We encode images into two codes: one that is common to both domains and one that is unique to the presence domain. Decoding from the common code yields healthy images; decoding with the addition of the unique code produces a residual change to this image that adds cancer. Translation proceeds from presence to absence and vice versa. In the first case, the tumor is re-added to the image and we successfully exploit the residual decoder to also perform segmentation. In the second case, unique codes are sampled, producing a distribution of possible tumors. To validate the method, we created challenging synthetic tasks and tumor segmentation datasets from public BRATS (brain, MRI) and LitS (liver, CT) datasets. We show a clear improvement (0.83 Dice on brain, 0.74 on liver) over baseline semi-supervised training with autoencoding (0.73, 0.66) and a mean teacher approach (0.75, 0.69), demonstrating the ability to generalize from smaller distributions of annotated samples.
% \dots
% \keywords{semi-supervised segmentation, weak labels, adversarial, image-to-image translation}
\end{abstract}
%%%%%%%%% KEYWORDS
%%%%%%%%% BODY TEXT
\section{Introduction}
\label{sec:introduction}
The segmentation of pathological anomalies in medical images can be performed efficiently with deep neural networks, but these require a large quantity of pixel-level annotations. Obtaining a sufficient quantity of annotations is difficult, costly, and sometimes impractical; on the other hand, unlabeled or weakly annotated data is easier to obtain. This weakly annotated data can be used to improve the performance of a semi-supervised segmentation model. Specifically, we target the scenario of medical images acquired for oncology purposes, where two image sets are available: \textit{(i)} those that are diagnosed as healthy and \textit{(ii)} those that are diagnosed as pathological but for which only some of the pathological images have pixel-level annotations. We use this weak image-level annotation to extend the training of the segmentation model to images that lack pixel-level annotations.

\begin{figure*}
    \centering
    \includegraphics[width=0.4\linewidth]{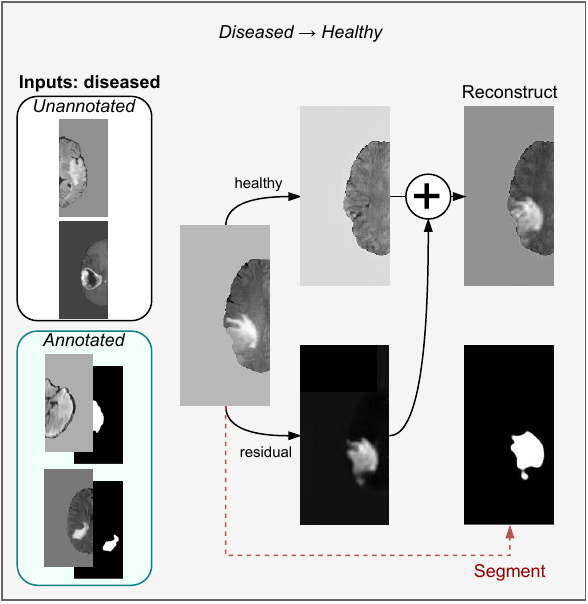}%0.47
    \includegraphics[width=0.4\linewidth]{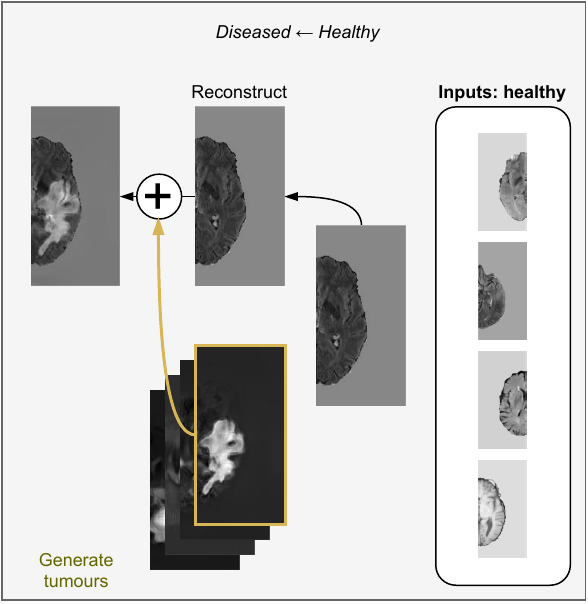}%0.47
    \caption{We address a common scenario in medical image segmentation where cases are known to either be diseased (eg. contains a tumor) or healthy (eg. no tumor) but only some of the diseased cases have pixel-level annotations for training an automated segmentation model. Slices of brain hemispheres from magnetic resonance imaging volumes are used for illustration purposes. We propose a model that uses image-to-image translation, from diseased to healthy and from healthy to diseased, in order to use the unannotated data to help it learn the segmentation task. \textit{Left:} When transforming a diseased image into a healthy image, a residual image output locates the tumor. This output is similar to the segmentation output (dashed red line). Summing the residual image with the generated healthy image recreates the original diseased image. \textit{Right:} When transforming a healthy image into a diseased image, residual images to add tumors need to be sampled from a learned prior distribution.}
    \label{fig:teaser}
\end{figure*}

The use of generative adversarial networks (GANs) has been explored to improve the semantic segmentation of medical images. Most of these improve segmentation with the addition of an adversarial objective ~\cite{kohl2017adversarial,dai2017scan,xue2018segan,moeskops2017adversarial,son2017retinal,zhu2016adversarial,yang2017automatic,rezaei2018whole,li2017brain}. Others perform data augmentation within the training set \cite{guibas2017synthetic,mok2018learning,izadi2018generative}; however, they do not augment the training set beyond annotated examples. On the other hand, some works have explored unsupervised anomaly localization using autoencoding \cite{baur2018deep} or GANs \cite{schlegl2017unsupervised,chen2018unsupervised} to learn a generative model of healthy cases. Another GAN-based approach is to train an error model that could be used for updates on unlabeled data \cite{zhang2017deep}. However, these approaches are approximate and do not make full use of available weak labels (healthy and diseased domain labels). Making better use of available data, some recent approaches relied on image-to-image translation between diseased and healthy cases \cite{baumgartner2017visual,andermatt2018pathology} but these were unsupervised and either approximate or not validated against baselines or on multiple tasks.

We focus on the common scenario of tumor segmentation where a large number of images lack pixel-level segmentation annotations but are weakly annotated in that they are known to be either healthy or diseased cases. This binary weak annotation identifies whether there is something to be segmented in an image. For example, when segmenting cancerous lesions, images marked ‘healthy’ do not contain cancerous tumors, while images marked ‘diseased’ do. We use a domain translation objective, transforming diseased images to healthy images (and vice versa) to reveal the boundaries of the pathological artifacts in the diseased images. This allows the segmentation model to use this weakly labeled data for training in addition to the diseased cases that have reference pixel-level annotations. We argue that the objective of translating from diseased to healthy images is similar to a segmentation objective. Consequently, we develop a semi-supervised segmentation method\footnote{Code for model and experiments: {[\textit{hidden during review period}]}} with image-to-image translation, trained on unpaired images from diseased and healthy domains.

Considering the healthy domain as a subset of the information in the diseased domain, we encode images into two latent codes: information that is \textit{common} to both and information that is \textit{unique} to the diseased domain. This allows decoding in two parts: (1) a `healthy' image decoder from the \textit{common} latent code and (2) a residual decoder from the \textit{unique} and \textit{common} codes, producing the residual (additive) change required to make the `healthy' image `diseased' (example in Fig.~\ref{fig:teaser}).

Because the the residual decoder produces the object that is to be segmented, its output is similar to a segmentation. Consequently, we re-use the decoder to also perform segmentation. In doing so, we maximize the proportion of model parameters that receive updates even during unsupervised training, when there are no pixel-level annotations available. Furthermore, whereas image-to-image translation models do not use long skip connections from the encoder to the decoder, we propose a long skip connection variant in our method. Long skip connections are common with supervised encoder-decoder models \cite{drozdzal2016importance}, where they help preserve spatial detail in the decoder even when the encoding is very deep.

In this paper, we propose a semi-supervised segmentation method that relies on image-to-image translation to use a small quantity of annotations efficiently. Specifically, we propose:
\begin{enumerate}
%[topsep=1pt,itemsep=1pt,partopsep=0pt, parsep=0pt,leftmargin=\labelwidth]%[noitemsep]
    \item New segmentation tasks (brain, liver, synthetic) with both `healthy' and `diseased' cases.
    \item A new semi-supervised segmentation method for these tasks, using adversarial image-to-image translation.
    \item Maximizing parameter updates by sharing a decoder for both translation and segmentation.
    \item New long skip connections.
\end{enumerate}
We validate our method on brain tumor in MR images, colorectal liver metastases in CT images, and challenging synthetic data, significantly improving over well-tuned baselines.

\section{Related works}
\vspace{1mm}
\noindent \textbf{Anomaly localization.} Generative models have been used to fit the distribution of healthy images in order to locate anomalies in images. To localize lesions in brain MRI that are known to be either healthy or with cancer, \cite{baur2018deep} fit the healthy data distribution with an autoencoder. Given an image presenting a lesion, the reconstruction of the image is likely to appear healthy, allowing rough localization of the lesion via the difference between the image and its reconstruction. Similarly, \cite{schlegl2017unsupervised} and \cite{chen2018unsupervised} employ a GAN to locate anomalies in retinal images and brain MR images, respectively. While these models require that weak 'diseased' or 'healthy' labels are known, they are trained only on the latter. Furthermore, they allow only rough unsupervised localization.

\vspace{1mm}
\noindent \textbf{Semi-supervised segmentation.} By translating from diseased to healthy images, \cite{baumgartner2017visual} trains a network to localize Alzheimer's derived brain morphological changes using the output residual. Bidirectional translation is used in \cite{andermatt2018pathology} via a multi-modal variant of CycleGAN \cite{zhu2017unpaired} applied to brain MRI with presence of tumors. Diseased images that are translated to healthy images are reverted back to the original image space via a residual inpainting of the lesions. Lesions are localized and segmented by predicting a minimal region to which to apply inpainting. Segmentation is unsupervised, while using a prior term minimizing the inpainting region. However it has been tested on a single dataset with no comparison to other unsupervised methods, and has not been extended to a weakly- or semi-supervised setting. Our work differs in that we propose a semi-supervised architecture that uses fewer parameters by reusing mappings, skipping information from the encoder to decoders, and proposing a decoder that is trained with both translation and segmentation objectives, where we validate the method on multiple tasks.

A semi-supervised segmentation method for medical images was proposed in \cite{zhang2017deep}, where a discriminator learns, on the annotated dataset, a segmentation corrector that can be used with unannotated data. This method may be limited in how well it could scale with the proportion of unannotated data since the discriminator's behaviour may not generalize well beyond the annotated dataset on which it is trained. Because this method can be applied to the output of any segmentation model, we consider it complementary to our proposed method. Similarly, the mean teacher method was adapted for segmentation \cite{perone2018deep}. In this formulation, the teacher network is an exponential moving average of weights from the student network and the student network seeks to learn to segment (supervised) and to be consistent with the teacher network’s predictions (unsupervised). As before, the reliability of the teacher may depend on the size of the training set.

\vspace{1mm}
\noindent \textbf{Adversarial learning.} Image to image translation is most prominently done with the CycleGAN \cite{zhu2017unpaired}, which performs bidirectional translation between two domains. UNIT \cite{liu2017unsupervised} proposed a similar approach but with a common latent space, shared by both domains, from which latent codes could be sampled. Augmented CycleGAN \cite{almahairi2018augmented} and Multimodal UNIT \cite{huang2018multimodal} respectively extended both methods from one-to-one mappings to many-to-many.

The one-to-many mapping required to translate healthy images into diseased images (there are many ways an image can appear diseased) is turned into a many-to-many mapping in~\cite{xia2019adversarial}. By jointly inputting a lesion segmentation mask with a healthy mask, a lesion is introduced to the image as specified in the mask.

Both \cite{huang2018multimodal} and \cite{lee2018diverse} present methods that learn shared and domain-specific latent codes to disentangle domain-specific variations. These differ from the proposed method in that they do not segment and do not assume (and benefit from) an \textit{absence} domain as a subset of a \textit{presence} domain. In addition, the domain-specific "style" codes are encoded with a shallow network which may bias the model to indeed learn domain-specific styles; whereas, the proposed method uses deep encodings (enabled by long skip connections from the encoder to the decoders) for all codes. Explicit disentangling of variations between these codes has recently been proposed in \cite{gonzalez2018image} by way of a gradient reversal layer \cite{ganin2015unsupervised}. Work by~\cite{kobayashi2020decomposing} that is concurrent to and related to ours uses image-to-image translation and segmentation objectives to learn normal and abnormal latent codes of medical images to aid in image retrieval.

Data augmentation is achieved with GANs for liver lesion classification in \cite{frid2018gan}. In \cite{jin2018ct}, data is synthesized to cover uncommon cases such as peripheral nodules touching the lung boundary. A segmentation mask generator is introduced in \cite{guibas2017synthetic} and \cite{mok2018learning} to augment small training datasets. Pathological x-ray images are generated from healthy images to augment the training dataset for segmentation in~\cite{tang2019xlsor}. Our method goes beyond data augmentation to achieve semi-supervised segmentation with GAN-based image-to-image translation employed in a way that disentangles the details needed for segmentation. 
\section{Methods}
\label{sec:method}

Segmentation labels are often available for an insufficiently representative sample of data. We propose a semi-supervised method that extends supervised segmentation to weakly labeled data using a domain translation objective. In addition to a segmentation objective, the method attempts to learn the translation between the distribution of images presenting the segmentation target (P) and the distribution of images where this target is absent (A). In section \ref{sec:method_objectives}, we argue that this is an appropriate choice of training objective for semi-supervised segmentation. Our training method is then detailed in \ref{sec:our_method} in a manner that is agnostic to the underlying network architectures. The baseline methods, network architectures, and training details are discussed in the Experiments section (\ref{sec:methods_baseline}, \ref{sec:architectures_and_training}).

\subsection{Translation, segmentation, and autoencoding}
\label{sec:method_objectives}

Translating images between two domains, one with a segmentation target present (\textit{presence domain} P) and another with is absent (\textit{absence domain} A) requires a model to learn segmentation implicitly. To remove the segmentation target, a model must learn to localize it and disentangle it from everything else in the image. We conjecture that segmentation relies on the same disentangling as translation and that this is the most difficult part of both objectives. Autoencoding of the model's input, the canonical objective of unsupervised feature learning, also disentangles variations in an image, as demonstrated by the winners of the Brain Tumor Segmentation challenge (2018) \cite{myronenko20183d} by augmenting a fully convolutional segmentation network with an autoencoding objective. However, diagnostic information about presence (P) or absence (A) of the segmentation target in the image may guide a domain translation objective to more specifically isolate the information that is important for segmentation \cite{gonzalez2018image}. Thus, we identify domain translation as a particularly useful objective for semi-supevised segmentation. For example, the tumor localized in Fig.~\ref{fig:teaser} is visible in the residual image which is similar to a tumor segmentation. We propose an encoder-decoder model leverages the similarity between the translation residual image and segmentation to employ a decoder that is shared by both objectives. Implicit data augmentation is achieved by translating not only from diseased to healthy but also vice versa. With our choice of objectives, we designed the method to scale better (in terms of segmentation performance) with the proportion of unannotated data than baseline methods.

\subsection{Semi-supervised segmentation via image-to-image translation}
\label{sec:our_method}

\begin{figure}[htb!]
    \centering
    \includegraphics[width=0.99\linewidth]{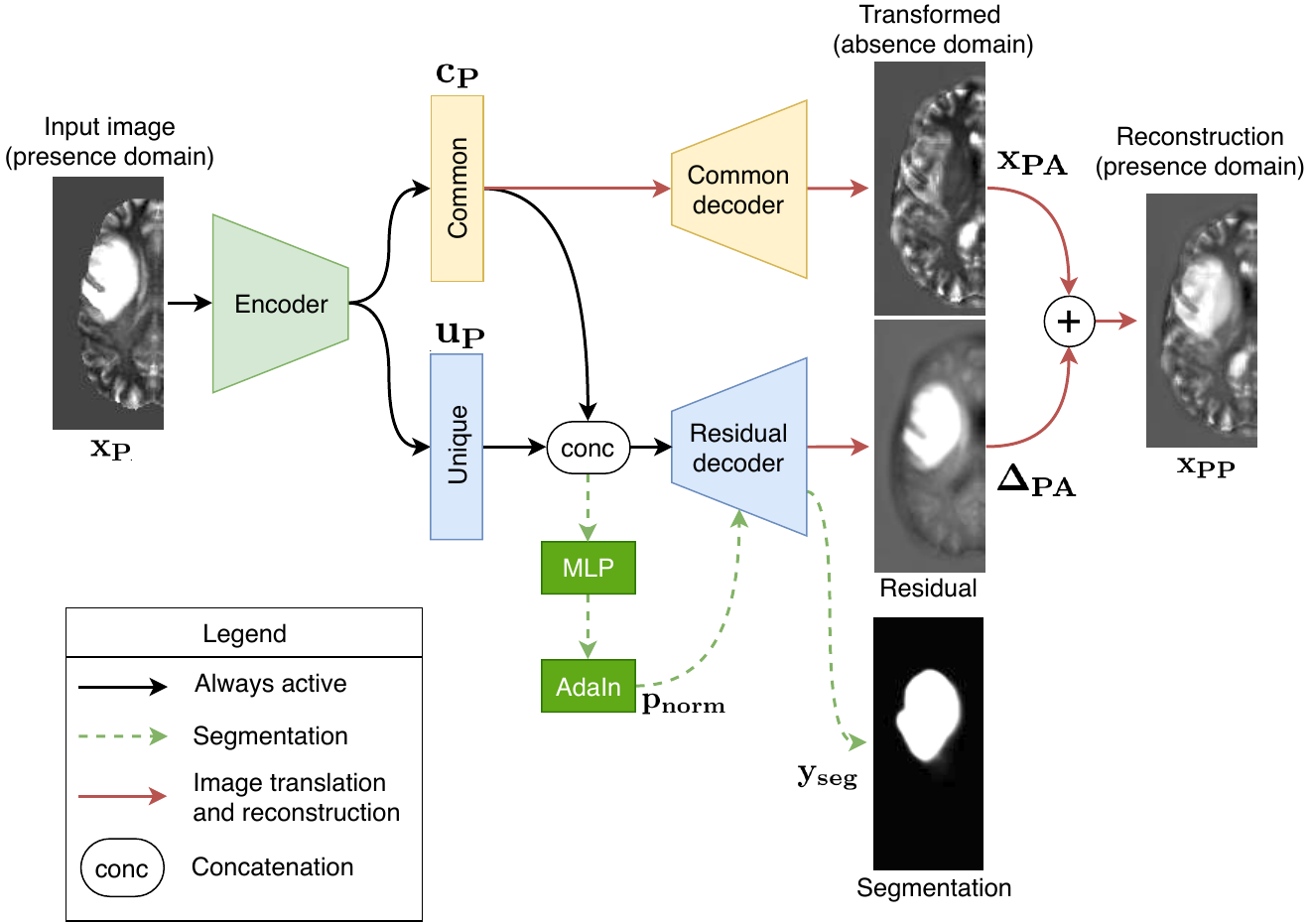}
    \caption{Simultaneous segmentation, image translation and reconstruction. Images are transformed from the \textit{presence} to the \textit{absence} domain (discriminator not shown).
    }
    \label{fig:method_pap}
    %%source at https://drive.google.com/file/d/1b7PEzTlUAwXc4N0giQBjicoVIe0sGwkE/view?usp=sharing
\end{figure}

\begin{figure}[htb!]
    \centering
    \includegraphics[width=0.99\linewidth]{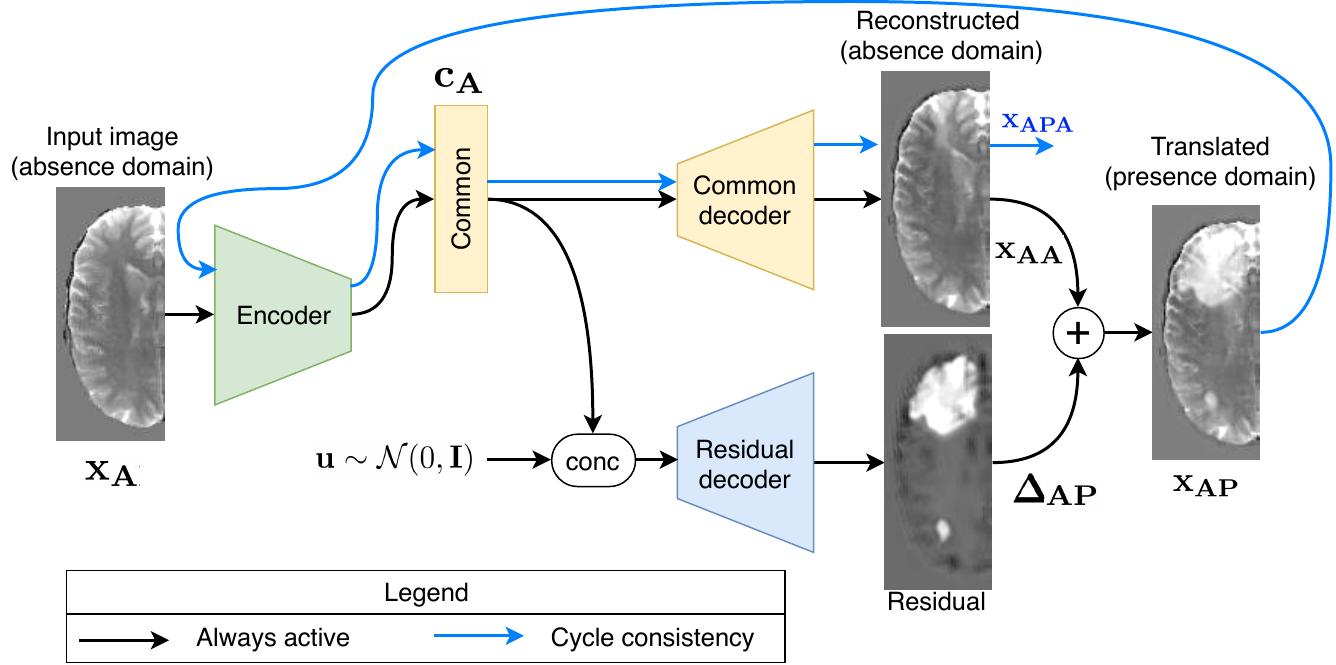}
    \caption{Image translation from the \textit{absence} to the \textit{presence} domain (discriminator not shown). The input image is reconstructed from the common code. An image in the \textit{presence} domain is generated by adding a residual change to this reconstruction. Blue lines: cycle consistency path.}
    \label{fig:method_apa}
\end{figure}

The proposed model builds on an encoder-decoder fully convolutional network (FCN) segmentation setup by introducing translation between a domain of images $\mathbf{x}$ presenting the segmentation target (P) and a domain where it is absent (A), $\mathbf{x_P}$ to $\mathbf{x_{PA}}$, as shown in Fig.~\ref{fig:method_pap}. The encoder separates codes into those that are \textit{common} to both A and P and those that are \textit{unique} to P; essentially, the information in A is a subset of the information in P. For example, in the case of medical images with tumors, both A and P contain the same organs but P additionally contains cancerous tumors. 

\textbf{Latent code decomposition.} Starting with images $\mathbf{x}$ in domains A or P ($\mathbf{x_A}$ or $\mathbf{x_P}$), the encoder ($\mathit{f}$) yields \textit{common} ($\mathbf{c}$) and \textit{unique} ($\mathbf{u}$) codes:

\begin{equation}
\label{eq:decomposition}
\begin{aligned}
\left[\mathbf{c_A, u_A}\right] &= \mathit{f}(\mathbf{x_A}),\\
\left[\mathbf{c_P, u_P}\right] &= \mathit{f}(\mathbf{x_P}).
\end{aligned}
\end{equation}

This decomposition of the latent codes is reminiscent of the \textit{style} and \textit{content} decomposition in \cite{huang2018multimodal} or the domain-specific codes in \cite{lee2018diverse}.

\textbf{Presence to absence translation.} Translation is achieved by selectively decoding from the latent codes $\mathbf{c}$ and $\mathbf{u}$. A \textit{common} decoder ($\mathit{g_{com}}$) uses only the common code, $\mathbf{c}$, to generate images in A:
\begin{equation}
\label{eq:common_decoder}
\begin{aligned}
\mathbf{x_{AA}} &= \mathit{g_{com}}(\mathbf{c_A}),\\
\mathbf{x_{PA}} &= \mathit{g_{com}}(\mathbf{c_P}),
\end{aligned}
\end{equation}
where $\mathbf{x_{AA}}$ is essentially an autoencoding of $\mathbf{x_A}$, whereas $\mathbf{x_{PA}}$ is an image translation of $\mathbf{x_P}$ from the P domain to the A domain where the segmentation target is removed. With this translation, the target can be recovered separately, by computing a residual change $\mathbf{\Delta_{PA}}$ to $\mathbf{x_{PA}}$ that reconstructs $\mathbf{x_{P}}$ as $\mathbf{x_{PP}}$. The residual change is computed by a \textit{residual} decoder ($\mathit{g_{res}}$) which uses all codes: those extracted from $\mathbf{x_P}$ that are common to both domains ($\mathbf{c_P}$) together with those that are unique to P ($\mathbf{u_P}$) as in  Fig.~\ref{fig:method_pap}:
\begin{equation}
\label{eq:residual}
\begin{aligned}
\mathbf{x_{PP}} &= \mathbf{x_{PA}}+\mathbf{\Delta_{PA}},\\
\textrm{where}~\mathbf{\Delta_{PA}} &= \mathit{g_{res}}(\mathbf{c_P}, \mathbf{u_P}).
\end{aligned}
\end{equation}

The residual decoder ($\mathit{g_{res}}$) takes all of the latent codes because the unique code decodes to the image space in a way that is dependent on the common code. For example, the way cancer manifests in a brain or liver scan depends on the location and structure of the brain or liver in the scan. Note also that because the common decoder ($\mathit{g_{com}}$, Eq. (\ref{eq:common_decoder})) only uses the common latent code, the encoder must learn to disentangle common and unique codes.

\textbf{Segmentation.} The $\mathbf{c_P}$ and $\mathbf{u_P}$ codes contain sufficient information for segmentation. Indeed we reuse the residual decoder, used with $\mathbf{x_P}$, for segmentation. We parameterize a segmentation decoder $\mathit{g_{seg}}$ using the residual decoder $\mathit{g_{res}}$ to produce a segmentation $\mathbf{y}$:
\begin{equation}
\label{eq:segmentation}
\begin{aligned}
\mathbf{y} &= \mathit{g_{seg}}(\mathbf{c_P}, \mathbf{u_P}),\\
&= (\mathit{\hat g_{res}} \circ \mathit{s}) (\mathbf{c_P}, \mathbf{u_P}),
\end{aligned}
\end{equation}
where $\mathit{s}$ is a pixel classification layer and $\mathit{\hat g_{res}}$ is a subset of the $\mathit{g_{res}}$ network that contains all but the last layer. Normalization parameters in $\mathit{g_{res}}$ are replaced by ones specific to the segmentation task, allowing $\mathit{g_{seg}}$ to produce outputs that differ from $\mathit{\hat g_{res}}$.

\textbf{Absence to presence translation.} Finally, we conclude the set of autoencoding and translation equations with $\mathbf{x_{AP}}$ and $\mathbf{x_{APA}}$, where images in A are translated to images in P (Fig.~\ref{fig:method_apa}). We note that although these translations are not useful for segmentation, they are useful during training since they effectively augment the training updates that our encoders and decoders can receive. Since P contains additional information to that found in A, we sample a random code from a prior distribution:
\begin{equation}
\label{eq:reverse_translation}
\begin{aligned}
\mathbf{x_{AP}} &= \mathit{g_{com}}(\mathbf{c_A})
                  +\mathit{g_{res}}(\mathbf{c_A}, \mathbf{u} \sim \mathcal{N}(0, \mathbf{I})),\\
\mathbf{x_{APA}} &= \mathit{g_{com}}(\mathbf{c_{AP}}),\\
\left[\mathbf{c_{AP}, u_{AP}}\right] &= \mathit{f}(\mathbf{x_{AP}}).
\end{aligned}
\end{equation}
Here, $\mathbf{x_{AP}}$ is the image translation of $\mathbf{x_A}$ in A to domain P using the common decoder $\mathit{g_{com}}$ for the code $\mathbf{c_A}$ that is common to both domains and the residual decoder $\mathit{g_{res}}$ for the code $\mathbf{u}$ that is unique to P; since this unique information is absent from A, it is randomly sampled from a zero-mean, unit variance Normal prior distribution, $\mathcal{N}(0, \mathbf{I})$. Note that unlike in a variational autoencoder, the encoder $\mathit{f}$ (Eq. (\ref{eq:common_decoder})) does not parameterize a conditional distribution over the unique codes but rather encodes a sample directly. We ensure that the distribution of encoded samples matches the prior by making $\mathbf{u_{AP}}$ match $\mathbf{u}$, where $\mathbf{u_{AP}}$ is obtained by encoding the image $\mathbf{x_{AP}}$ that is produced using the random sample $\mathbf{u}$. The translation of $\mathbf{x_{AP}}$ to $\mathbf{x_{APA}}$ completes a cycle as in \cite{zhu2017unpaired}. When $\mathbf{x_{APA}}$ must match $\mathbf{x_A}$, this ensures that the translations retain information about their source images, ensuring that the encoder and decoders do not learn trivial functions. As shall be seen below, this is already achieved by other objectives, making the cycle optional.

\textbf{Total loss.} The training objective consists of a segmentation loss $\mathit{L_{seg}}$ combined with four translation losses, each weighted by some scalar $\mathit{\lambda}$: 
\begin{equation}
\begin{aligned}
\mathit{L_{total}} = \mathit{L_{seg}} &+ \mathit{\lambda_{rec} L_{rec}} + \mathit{\lambda_{lat} L_{lat}} \\
&+ \mathit{\lambda_{cyc} L_{cyc}} + \mathit{\lambda_{adv} L_{adv}}.
\end{aligned}
\end{equation}

\textbf{Segmentation loss.} We use a soft, differentiable Dice loss for segmentation, as in \cite{drozdzal2016importance,milletari2016v}, which measures the overlap between the predicted segmentation $\mathbf{y}$ (Eq. (\ref{eq:segmentation})) and reference segmentation $\mathbf{\hat y}$:
\begin{equation}
\mathit{L_{seg}} = \text{Dice}(\mathbf{y}, \mathbf{\hat y}).
\end{equation}

\textbf{Reconstruction losses.} To ensure that the encoder and decoders can cover the distribution of images, we reconstruct input images (see Eqs. (\ref{eq:decomposition}), (\ref{eq:residual})):
\begin{equation}
\mathit{L_{rec}} = \mathit{L_{rec}}(\mathbf{x_P}, \mathbf{x_{PP}}) + \mathit{L_{rec}}(\mathbf{x_A}, \mathbf{x_{AA}}).
\end{equation}

Similarly, we reconstruct the latent codes of translated images $\mathbf{x_{AP}}$, $\mathbf{x_{PA}}$ and reconstructed images $\mathbf{x_{AA}}$, $\mathbf{x_{PP}}$ by re-using the encoder $\mathit{f}$ so as to ensure that their distributions match across domains A and P
\begin{equation}
\begin{aligned}
\left[\mathbf{c_{AP}, u_{AP}}\right] &= \mathit{f}(\mathbf{x_{AP}})\\
\left[\mathbf{c_{PA}, u_{PA}}\right] &= \mathit{f}(\mathbf{x_{PA}})\\
\left[\mathbf{c_{AA}, u_{AA}}\right] &= \mathit{f}(\mathbf{x_{AA}})\\
\left[\mathbf{c_{PP}, u_{PP}}\right] &= \mathit{f}(\mathbf{x_{PP}})\\
\mathit{L_{lat}} &= 
     \mathit{L_{lat}}(\mathbf{c_A}, \mathbf{c_{AP}})
   + \mathit{L_{lat}}(\mathbf{c_P}, \mathbf{c_{PA}})
\\&+ \mathit{L_{lat}}(\mathbf{c_A}, \mathbf{c_{AA}})
   + \mathit{L_{lat}}(\mathbf{c_P}, \mathbf{c_{PP}})
\\&+ \mathit{L_{lat}}(\mathbf{u_P}, \mathbf{u_{PP}})
   + \mathit{L_{lat}}(\mathbf{u}, \mathbf{u_{AP}}),
\end{aligned}
\end{equation}
where $\mathbf{u}$ is a sample from the prior distribution, $\mathcal{N}(0, \mathbf{I}))$, and $\mathbf{c_A}$, $\mathbf{u_A}$, $\mathbf{c_P}$, and $\mathbf{u_P}$ are the original codes from Eq. (\ref{eq:decomposition}). This encourages the unique codes to match the prior.

The reconstruction objectives encourage the mappings to be bijective, ensuring that there is a unique output for every input. If an image or a latent space can be reconstructed, this means that the mapping does not discard information about it. Without the reconstruction objectives, the healthy parts of an input image may not be retained in the output and mode dropping may occur in image-to-image translation where many inputs map to the same output. This is because a single output that is independent of the choice of input is sufficient to satisfy a domain translation objective.

Furthermore, we define a cycle consistency loss for the APA cycle (Eq. (\ref{eq:reverse_translation})):
\begin{equation}
\mathit{L_{cyc}} = \mathit{L_{rec}}(\mathbf{x_A}, \mathbf{x_{APA}}).
\end{equation}
It should be noted that there is no PAP cycle since in the proposed method this is equivalent to PP reconstruction, as can be seen in Fig.~\ref{fig:method_pap}. Because both images and their latent codes are reconstructed, the cycle consistency loss is optional.

We use the $L_1$ distance for all reconstruction losses.

\textbf{Adversarial loss.} Finally, we use the hinge loss for the adversarial objective, together with spectral norm on the encoder and decoders as in \cite{zhang2018self}:
\begin{equation}
\begin{aligned}
\mathit{L_{adv}} = \sum_{d \in \{A, P\}} &\min_{\mathit{G}} \max_{\mathit{D}} \Big[\\
&-\mathbb{E}_{\mathbf{x_d} \sim \mathit{p_d}} \left[\text{min}(0, \mathit{D_d}(\mathbf{x_d})-1)\right]\\
&-\mathbb{E}_{\mathbf{\hat x_d} \sim \mathit{\hat p_d}} \left[\text{min}(0, -\mathit{D_d}(\mathit{G_d}(\mathbf{\hat x_d}))-1)\right]\\
&-\mathbb{E}_{\mathbf{\hat x_d} \sim \mathit{\hat p_d}} \mathit{D_d}(\mathit{G_d}(\mathbf{\hat x_d})) \Big],
\end{aligned}
\end{equation}
where, for each domain $\mathit{d} \in \{A, P\}$, $\mathit{G_d}$ is the generator network for some generated image $\mathbf{\hat x_d} \sim \mathit{\hat p_d}$ (specifically, $\mathbf{x_{AP}}$ or $\mathbf{x_{PA}}$) and $\mathit{D_d}$ is a discriminator network which discriminates between real data $\mathbf{x_d} \sim \mathit{p_d}$ and generated data $\mathbf{\hat x_d}$, effectively matching the distribution of generated images $\mathit{\hat p_d}$ to the distribution of real images $\mathit{p_d}$.

\subsection{Compressed long skip connections}
\label{sec:long_skip}

In the proposed and baseline methods, every decoder (except for the reconstruction decoder in the autoencoding baseline) accepts long skip connections from the encoder, as in \cite{drozdzal2016importance}. These connections skip features from each layer in the encoder to the corresponding layer in the decoder, except for the first and the last layers. Because long skip connections make autoencoding trivial, they are used only with the segmentation decoder of the autoencoding baseline and not with its reconstruction decoder.
%\evnote{highlight that other translation methods do not use long skip connections}

% \begin{figure}[htb!]
% \centering
% \includegraphics[width=0.99\linewidth,trim = 22cm 1cm 0 0,clip]{images/skinny_cat.pdf}
% \caption{Compressed skip connection as a way to limit information skipped forward while preserving spatial cues.}
% \label{fig:skinny_cat}
% % source at https://drive.google.com/file/d/1b7PEzTlUAwXc4N0giQBjicoVIe0sGwkE/view?usp=sharing
% \end{figure}

Typically, feature maps from the encoder are either directly summed with \cite{drozdzal2016importance} or concatenated to \cite{ronneberger2015u} those in the decoder. We propose here a modified variant of long skip connections where any stack of feature maps is first compressed (via $1\times1$ convolution) to a single map before concatenation. We note that concatenating all feature maps is costly computationally and appears to increase training time for image translation whereas summing feature maps makes the image translation task difficult to learn. To further stabilize training, all features skipped from the encoder are normalized with instance normalization~\cite{ulyanov2016instance}. We find that these long skip connections help train the model faster and help produce higher quality image outputs even with a deep encoder.

\subsection{Shared network for translation and segmentation}

Finally, in the proposed method, the \textit{residual} decoder network (Eq. (\ref{eq:residual})) is used both for translation and for segmentation. For segmentation, all but the last layer is replaced by a classification layer (1$\times$1 convolution with 1 output channel). The segmentation output differs from the translation output both in texture and global structure. For example, the translation output for brain tumor segmentation produces a tumor but also changes to the surrounding brain. Therefore, to produce a segmentation, the representation in the \textit{residual} decoder is modified at every scale: a different normalization is applied to each layer depending on whether the output is a translation residual or a segmentation.

% %%%%%%%%% Experiments
\section{Experiments}
\label{sec:experiments}

We evaluate our proposed semi-supervised segmentation method on both synthetic and real data (brain and liver images). First, we demonstrate our method on challenging synthetic data, then we validate it on brain tumor segmentation in MRI and on liver tumor segmentation in CT. We present ablation studies in the end.

\subsection{Baseline methods}
\label{sec:methods_baseline}
The proposed method is compared against three baseline approaches: (1) a fully supervised encoder-decoder model (``Segmentation only''); (2) the same but semi-supervised with an additional decoder with an autoencoding objective (``AE baseline''); and (3) a semi-supervised mean teacher approach~\cite{perone2018deep} (``Mean teacher''). To allow for direct comparison, all models (baselines and proposed) share the same encoder and decoder architectures (detailed in Table~\ref{tab:arch}). To better match the proposed method, the autoencoding baseline reconstructs images from both domains A (absence, healthy) and P (presence, diseased). All encoder-decoder models are inspired by the UNet~\cite{ronneberger2015u} which employs a convolutional neural network (CNN) to encode features at decreasing resolution and increasing field of view (encoder) and an upsampling CNN that decodes the features into a full resolution output (decoder). The proposed method has one encoder and two decoders (\textit{common} and \textit{residual}). The fully supervised and mean teacher baselines use a single encoder with a single decoder. This is equivalent to the proposed method with only the segmentation loss, using only the residual decoder. The AE baseline uses one encoder and two decoders; the additional decoder reconstructs the input.

\subsection{Network architectures and training}
\label{sec:architectures_and_training}
To compare the effect of different training objectives, we try to reduce the confounding effect of differing architectures between the proposed model and baseline models. For each task, we use the same encoder ($\mathit{f}$, Eq. (\ref{eq:decomposition})) for all models; likewise, the \textit{common} decoder (Eq. (\ref{eq:common_decoder})) in the proposed model and all decoders in the baseline models are the same. The \textit{residual} decoder in the proposed model is similar, differing in that it lacks short skip connections and uses slightly larger convolution kernels.

\textbf{Encoder and decoder architectures.}
Network layers are denoted in Table~\ref{tab:arch}. For 48$\times$48 synthetic tasks, the encoder is missing the first NRC block (starts with C 32) and the decoders are missing the last. The discriminator for all synthetic tasks lacks the first NRC block (starts with C 128). All decoders except those used for input reconstruction in the autoencoder baseline models receive long skip connections from the encoder of the type proposed in section~\ref{sec:long_skip}. All latent bottleneck representations of every model have 512 channels. In the proposed model, 128 of these channels are specified as the \textit{unique} latent code and the rest as the \textit{common} latent code. Mean teacher models~\cite{perone2018deep} were trained with an exponential moving average weight of 0.99 (applied constantly, with no gradual change as this worked better on our data) and consistency weight of 0.01 (selected from \{0.01, 1, 1, 10\}). The best mean teacher model variation was also selected on the validation set for each task: using dropout but no normalization layers, using normalization layers but no dropout, or using both dropout and normalization layers. All encoders and decoders were initialized with the Kaiming Normal approach~\cite{he2015delving}. Convolutions are applied to inputs with reflection padding. For all experiments we use PyTorch~\cite{pytorch}.

\begin{table*}[h]
\caption{Architecture used for the proposed and baseline methods. `Ch' is the number of channels. The decoder refers to the \textit{common} decoder in the proposed method and all decoders in the baseline methods. C $=$ convolution, NR $=$ normalization $+$ ReLU nonlinearity, NRC $=$ normalization $+$ ReLU $+$ convolution; * = a short skip connection.}
\label{tab:arch}
\centering
\setlength\tabcolsep{4pt}
\begin{subtable}[t]{0.23\textwidth}
\begin{tabular}{ cccc }
  \multicolumn{4}{c}{\textbf{Encoder}} \\
  \hline\hline
  Layer & Ch & Kernel & Stride \\
  \hline
  C & 16 & 3$\times$3 & 1 \\
  NRC* & 32 & 3$\times$3 & 2 \\
  NRC* & 64 & 3$\times$3 & 2 \\
  NRC* & 128 & 3$\times$3 & 2 \\
  NRC* & 256  & 3$\times$3 & 2 \\
  NRC* & 512 & 3$\times$3 & 2 \\
  NR & & & \\
\end{tabular}
\end{subtable}
\qquad
\begin{subtable}[t]{0.23\textwidth}
\begin{tabular}{ cccc }
  \multicolumn{4}{c}{\textbf{Decoder}} \\
  \hline\hline
  Layer & Ch & Kernel & Stride \\
  \hline
  C & 256 & 3$\times$3 & 1 \\
  NRC* & 128 & 3$\times$3 & 1 \\
  NRC* & 64 & 3$\times$3 & 1 \\
  NRC* & 32 & 3$\times$3 & 1 \\
  NRC* & 16 & 3$\times$3 & 1 \\
  NRC & 1 & 3$\times$3 & 1 \\
  & & & \\
\end{tabular}
\end{subtable}
\qquad
\begin{subtable}[t]{0.23\textwidth}
\begin{tabular}{ cccc }
  \multicolumn{4}{c}{\textbf{Residual decoder}} \\
  \hline\hline
  Layer & Ch & Kernel & Stride \\
  \hline
  C & 256 & 5$\times$5 & 1 \\
  NRC & 128 & 5$\times$5 & 1 \\
  NRC & 64 & 5$\times$5 & 1 \\
  NRC & 32 & 5$\times$5 & 1 \\
  NRC & 16 & 5$\times$5 & 1 \\
  NRC & 1 & 5$\times$5 & 1 \\
  & & & \\
\end{tabular}
\end{subtable}
\begin{subtable}[t]{0.23\textwidth}
\begin{tabular}{ cccc }
  \multicolumn{4}{c}{\textbf{Discriminator}} \\
  \hline\hline
  Layer & Ch & Kernel & Stride \\
  \hline
  C & 64 & 4$\times$4 & 1 \\
  NRC & 64 & 4$\times$4 & 2 \\
  NRC & 128 & 4$\times$4 & 2 \\
  NRC & 256 & 4$\times$4 & 2 \\
  NRC & 512 & 4$\times$4 & 2 \\
  C & 1 & 1$\times$1 & 1 \\
  & & & \\
\end{tabular}
\end{subtable}
\end{table*}

\textbf{Normalization.} All encoders use instance normalization~\cite{ulyanov2016instance}, while all decoders use layer normalization~\cite{ba2016layer}. In order to use the residual decoder for both translation and segmentation, its normalization layers use a separate set of parameters for each of these two tasks, allowing the same learned features to be used in different ways.

\textbf{Adversarial discriminators.} We use multi-scale discriminators as proposed in~\cite{wang2018high,liu2017unsupervised}. The discriminator architecture shown in Table~\ref{tab:arch} describes the network that is applied at each of the three scales, where the discriminator outputs a map of values per image instead of a single value. Discriminator outputs are averaged across all scales. All discriminators use leaky ReLU~\cite{maas2013rectifier} with a slope of $0.2$ and were initialized with the Kaiming Normal approach~\cite{he2015delving}.

\textbf{Optimization.} For all experiments, we used the AMSGrad optimizer~\cite{reddi2018convergence} with $\beta_1 = 0.5$ and $\beta_2 = 0.999$, a learning rate of $0.0001$, and batch sizes of 20 for brain and synthetic data and 10 for liver. Training was run for 500, 200, and 300 epochs for the brain, liver, and synthetic data, respectively. In the proposed method, we used the hinge loss for the adversarial objective, with spectral normalization~\cite{miyato2018spectral} applied to all networks, as in~\cite{zhang2018self,brock2018large}. All objective weights for the proposed method were always normalized so that they sum to 1; as such, the weights reported here are \textit{relative} weights. We found that the following objective weights yielded the best overall performance for brain and synthetic data when reference annotations were used for 1\% of the data: $\lambda_{avd}=3$, $\lambda_{rec}=50$, $\lambda_{lat}=1$, $\lambda_{cyc}=50$, $\lambda_{seg}=0.01$. (AE: $\lambda_{rec}=\lambda_{seg}=1$). For liver data, higher reconstruction weights were found to be sometimes required to avoid mode dropping in image-to-image translation; $\lambda_{rec}=200$ and $\lambda_{cyc}=200$ and/or $\lambda_{lat}=5$ were used whenever these produced a better score on the validation subset. When using reference annotations for 10\%, 20\%, and 100\% of the data, we set $\lambda_{dice}$ for the proposed method as $10$, $100$, and $1000$, respectively. Importantly, for brain and liver data, reference annotations were only made available for all or none of the slices from any single volume. This limits the variety of labeled cases in the training set.

\textbf{Data augmentation.} We applied data augmentation on the fly during training for brain and liver but not for synthetic tasks since a large amount of data is generated for the latter. Data augmentation involved random rotations up to 3 degrees, random zooms up to 10\% in or out, random intensity shifts up to 10\%, random horizontal and/or vertical flips, and spline warping. Spline warping used a 3$\times$3 grid of control points with each point placed according to a Normal distribution with variance $\sigma=5$. In those cases where data augmentation created new pixels along image edges or corners, these were filled by the reflection of the original image outward toward the edges and corners.

\textbf{Data batches}. Every batch of the training data, with a size $\mathit{N}$ contains $\mathit{N}$ diseased slices and $\mathit{N}$ unpaired healthy slices. We consider an epoch of training complete when every diseased slice has been provided to the model once. For the supervised segmentation objective, reference annotations are provided for some fraction (typically 1\%) of the diseased slices. For any fraction, the same set of diseased and healthy cases is used for all semi-supervised methods. For the fully supervised segmentation baseline, only the annotated subset of diseased slices is used and every batch contains only these slices.

\textbf{Multiple runs.} All experiments were run three times, with a different random model initialization each time, using three random seeds. Performance is reported as the mean score with a standard deviation of the means across the three runs. The variance across inputs is not considered.

\subsection{Synthetic tasks}

To simulate both simple and challenging examples in the data setup which is addressed with our method, we constructed an image set for synthetic segmentation tasks where some data is known to present the object of interest for segmentation (domain P), some data is known to be absent of the object of interest (domain A), and most of the data in P lacks reference segmentations. We constructed a synthetic task for digit segmentation using MNIST digits, similar to the cluttered MNIST dataset in \cite{jaderberg2015spatial}. Each image in P contains a complete randomly positioned digit placed on a background of clutter. The clutter is produced from random crops of digits sampled from within the same data fold (training, validation, or test set). In all experiments, we used crops of 10x10 pixels. Using the 60,000 MNIST training images, the same number of cluttered images were produced in P and in A, using each digit once in the foreground of images in P. 500 of these were used as a validation set for hyperparameter tuning. The 10,000 MNIST testing images were similarly used to produce the test set.

\begin{figure}[htb!]
    \centering
    \includegraphics[width=0.7\linewidth]{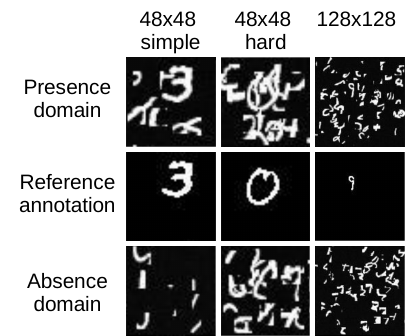}
    \caption{Examples of images from the synthetic datasets.
    }
    \label{fig:mnist_dataset}
\end{figure}

We tested the proposed model and the baseline methods on three variants of the synthetic task, at two resolutions: \textit{48$\times$48 simple}, with 8 pieces of clutter; \textit{48$\times$48 hard}, with 24 pieces of clutter; and \textit{128$\times$128}, with 80 pieces of clutter. Samples from these generated datasets are shown in Fig.~\ref{fig:mnist_dataset}; all datasets were generated prior to training. In all experiments, we provided reference segmentations for 1\% of the training examples. In addition, to mimic the issue of small training datasets where the training sample of images fails to cover all modes of variation of the population of images, we trained on reference segmentations only for the digit 9.

As shown in Table \ref{tab:results}, the proposed method clearly outperforms all baseline methods, achieving a Dice score of $0.84$, $0.62$, and $0.74$ on the simple, hard, and large variants of the task, respectively. The mean teacher baseline method nearly matches the proposed model ($0.83$ Dice) on the simple task and performs well on the other synthetic tasks. This suggests that the mean teacher model implementation that we use is competitive but struggles with more complex cases such as real brain and liver data. Overall, we conjecture that this synthetic data is particularly simple, compared to real medical data, as in the brain and liver tumour segmentation tasks where the proposed method clearly outperforms all baseline methods.

\begin{figure}[htb!]
    \centering
    \includegraphics[width=0.95\linewidth]{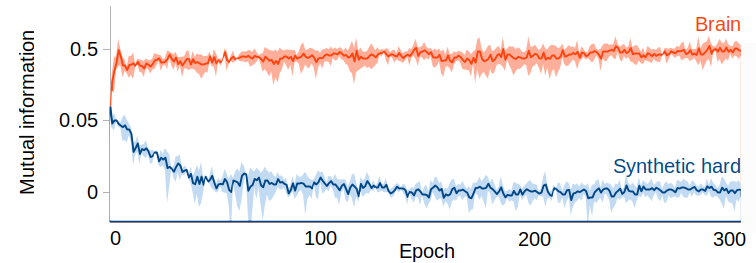}
    \caption{Mutual information computed between the common and unique codes, per epoch, for synthetic (48$\times$48 hard) and brain tumour segmentation tasks. Mean (solid) and standard deviation (shaded) across three training runs.
    }
    \label{fig:mutual_information}
\end{figure}

Specifically, the task of disentangling the segmentation target (digits) from the background (clutter) appears simple compared to real data and so may not benefit as much from our method's unsupervised ability to separate the target from the background. When a digit is placed over a background of clutter, it does not modify the background in any way. On the other hand, a tumour in a brain can affect the background by deforming all of the surrounding brain tissue. To verify whether synthetic data is disentangled better than real data, we evaluated the mutual information between the common and unique codes ($\mathbf{c_A}$ and $\mathbf{u_A}$ in Eq.~\ref{eq:decomposition}) throughout training on synthetic and on brain data by using a jointly trained Mutual Information Neural Estimator (MINE)~\cite{belghazi2018mine}. Indeed, we found that the model learns to completely disentangle the unique and common codes for synthetic data (nearly zero mutual information) but not for brain data, as shown in Figure~\ref{fig:mutual_information}.

\subsection{Brain tumor segmentation}

We created a 2D brain tumor segmentation dataset that corresponds to the problem definition for which we propose our segmentation training method. Specifically, we created a dataset with a set of known healthy cases (domain A) and a set of (diseased, domain P) cases known to present cancerous tumors. Pixel-level reference segmentations were made available for only a subset of the diseased cases. Both healthy and diseased cases were created from the 3D brain tumor segmentation challenge (BraTS) 2017 MRI volumes, all of which contain brain tumors.

Using 210 high grade glioma (HGG) volumes and 75 low grade glioma (LGG) volumes in the BraTS training set, we allocated 80\% of the data for training, 10\% of the data as a validation subset for hyperparameter tuning, model selection, and early stopping and 10\% for final testing and comparison to baseline methods. Training, validation, and testing subsets all had approximately equal proportions of HGG and LGG cases. No volumes were represented in more than one subset; models must generalize across cases. Splitting the volumes into axial slices as described below, 8475 healthy slices (domain A) and 7729 tumor slices (domain P) were generated.

Healthy cases were collected from axial slices without tumors. As in ~\cite{cohen2018distribution}, we split the slices into two hemispheres and selected only those half-slices of which at least 25\% of the area is brain tissue, so as to better balance the slice distributions between the two domains A and P. Considering that more tumors may be found in larger brain slices, this ensures that slice sizes and locations are more equally represented in A and P and the main difference between them is the absence or presence of tumor. In P, we also limited the minimal number of tumor pixels to 1\% of brain pixels.

We pre-processed every volume by mean-centering the brain regions and dividing them by five times their standard deviations, considering only brain tissue and ignoring the background. Finally, we used half-slices extracted from the processed volumes as model inputs. Each input has four channels, corresponding to four registered MRI sequences: T1, T2, T1C, and FLAIR.

\begin{table*}[htb!]
\caption{Segmentation Dice scores for the brain, liver, and synthetic data when annotations are available for only 1\% of the data: mean (standard deviation) across three runs.}
\centering
\setlength{\tabcolsep}{6pt}
% \begin{subtable}[t]{0.72\textwidth}
\begin{tabular}[t]{lccccc }
  %\toprule
  & Brain & Liver & Synthetic & Synthetic & Synthetic \\
  & 240$\times$120 & 240$\times$240 & 48$\times$48 simple & 48$\times$48 hard & 128$\times$128 \\
  \midrule
  Only segmentation & 0.69 (0.04) & 0.66 (0.01) & 0.61 (0.01) & 0.36 (0.01) & 0.15 (0.01) \\
  AE baseline & 0.73 (0.04) & 0.66 (0.02) & 0.75 (0.01) & 0.49 (0.02) & 0.57 (0.02) \\
  Mean teacher & 0.75 (0.01) & 0.69 (0.02) & 0.83 (0.02) & 0.56 (0.01) & 0.70 (0.01) \\
  Proposed & \textbf{0.83 (0.01)} & \textbf{0.74 (0.01)} & 0.\textbf{84 (0.00)} & 0.60 (0.01) & \textbf{0.74 (0.02)} \\
  Proposed (sep dec) & 0.79 (0.00) & 0.69 (0.02) & 0.83 (0.01) & \textbf{0.62 (0.01)} & 0.73 (0.02) \\
%   \bottomrule
\end{tabular}
% \caption{Proposed vs baselines}
\label{tab:results}
% \end{subtable}
% \quad
% \begin{subtable}[t]{0.22\textwidth}
% \begin{tabular}[t]{lc}
%   & Synthetic\\
%   & 48$\times$48 hard\\
%   \midrule
%   Compress & \textbf{0.57 (0.00)} \\ 
%   Concat & \textbf{0.57 (0.01)} \\
%   Sum & 0.56 (0.00) \\
%   No skip & 0.42 (0.01) \\
%   &\\
% \end{tabular}
% \caption{Ablations}
% \label{tab:ablations}
% \end{subtable}
\end{table*}

\begin{figure}
    \centering
    \includegraphics[width=0.6\linewidth]{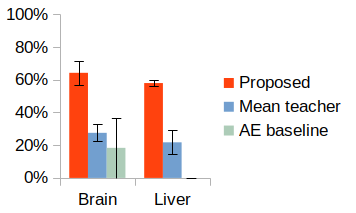}
    \caption{The performance gap, covered by each semi-supervised method, between supervised segmentation using 1\% of the data (lower bound) and all of the data (upper bound). At 0\%, a semi-supervised method performs as poorly as supervised segmentation trained on 1\% of the data; at 100\% it performs as well as a supervised method that is trained on all the data. (Error bars: standard deviation across three training runs).}
    \label{fig:relative}
\end{figure}

We trained the proposed model and baselines with reference annotations available for approximately 1\% of the training data. In order to sample the same number of healthy slices as diseased slices in each batch and epoch, some healthy slices were sampled more than once in an epoch. As shown in Table~\ref{tab:results}, the proposed model achieves a $0.83$ Dice score, significantly outperforming the segmentation baseline ($0.69$) and the semi-supervised autoencoding ($0.73$) and mean teacher ($0.75$) baselines. Figure~\ref{fig:relative} shows how much of the gap is covered by each method between segmentation using 1\% of the data vs using 100\% of the data; viewed in this way, the proposed method covers 64\% $\pm$ 7\% of the gap, as compared to 28\% $\pm$ 5\% or 18\% $\pm$ 18\% by the mean teacher and autoencoding baselines, respectively.

\begin{figure*}[t]
    \centering
    \includegraphics[width=0.056\linewidth]{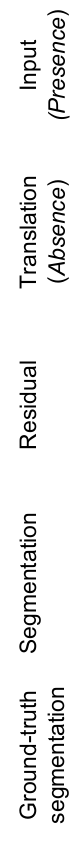}%0.07
    \includegraphics[width=0.064\linewidth]{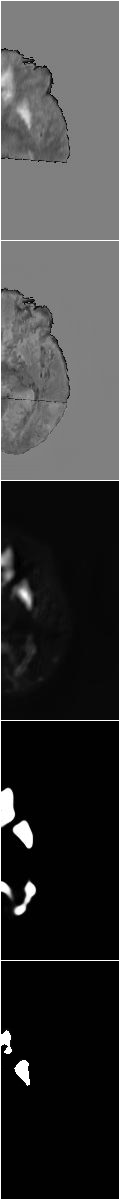}%flair, 0.08
    \includegraphics[width=0.064\linewidth]{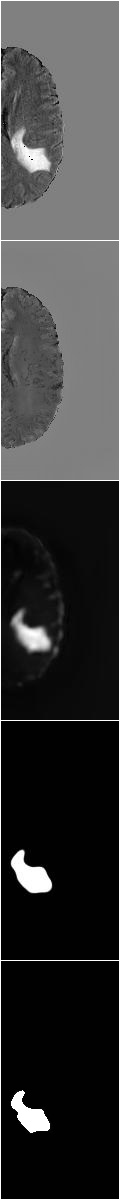}%flair, 0.08
    \includegraphics[width=0.064\linewidth]{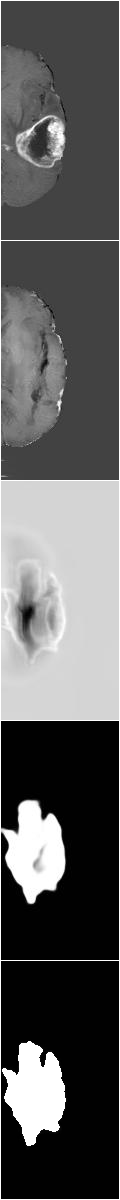}%t1, 0.08
    \includegraphics[width=0.064\linewidth]{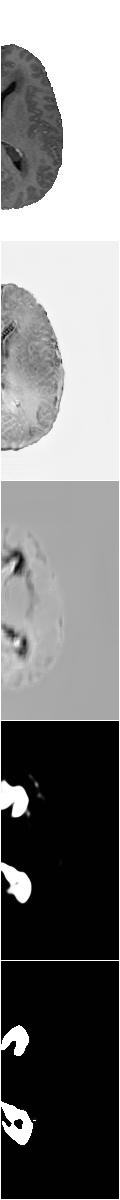}%t1, 0.08
    \includegraphics[width=0.064\linewidth]{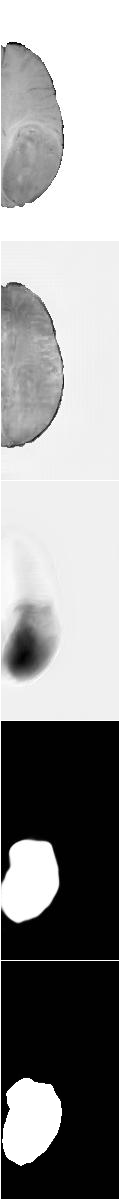}%t1c, 0.08
    \includegraphics[width=0.064\linewidth]{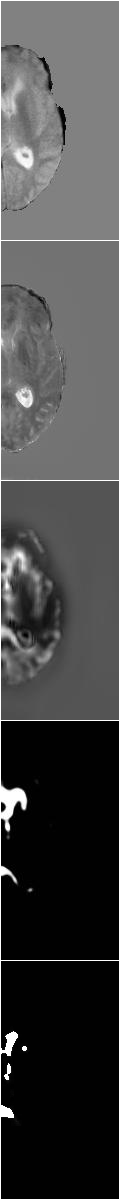}%t2, 0.08
    \includegraphics[width=0.064\linewidth]{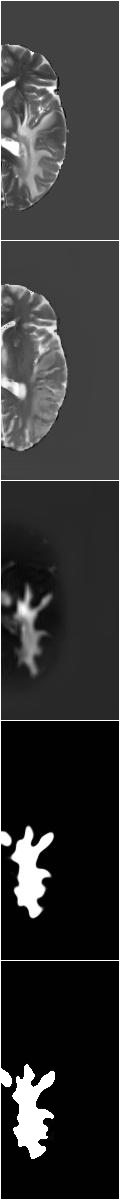}%t2, 0.08
    \quad\vline\quad
    \includegraphics[width=0.12584\linewidth]{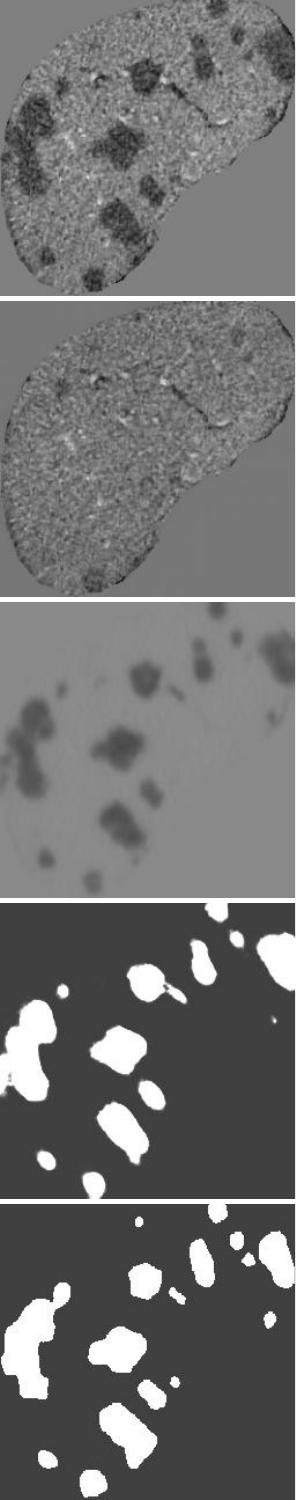}% 0.1573
    \
    \includegraphics[width=0.12584\linewidth]{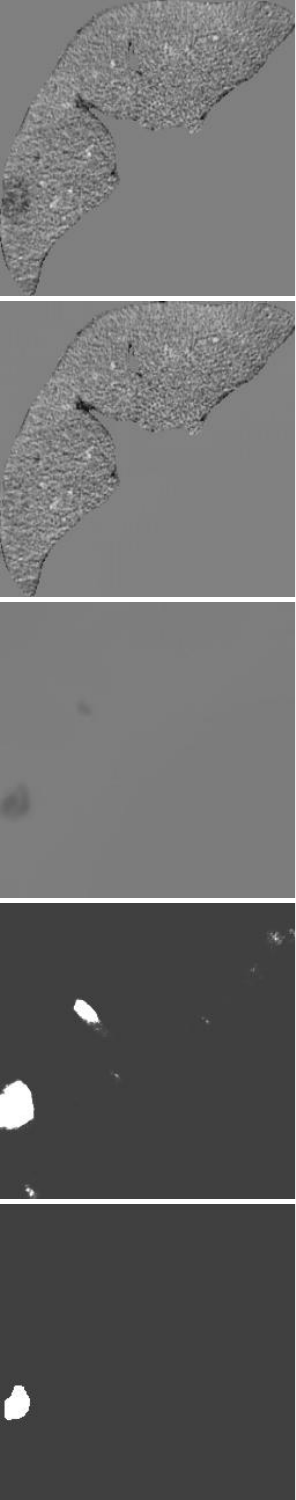}% 0.1573
    \\
    % \raggedright
    % \begin{tabular}{p{1cm}p{1cm}p{1cm}p{1cm}p{1cm}p{1cm}p{1cm}p{1cm}p{0.5cm}p{2.5cm}p{2.5cm}}
    \begin{tabular}{p{0.74cm}p{0.74cm}p{0.74cm}p{0.74cm}p{0.74cm}p{0.74cm}p{0.74cm}p{0.74cm}p{0.3cm}p{2cm}p{2cm}}
    % 1 & 2 & 3 & 4 & 5 & 6 & 7 & 8 & 9 & 10 & 11 \\
    & FLAIR & FLAIR & \hfil T1 & \hfil T1 & \hfil T1C & \hfil T2 & \hfil T2 & & \multicolumn{2}{c}{Liver} \\
    & \multicolumn{7}{c}{Brain} & & & \\
    \end{tabular}
    
    \caption{Examples, per column, of image segmentation and translation from \textit{Presence} to \textit{Absence} domains for brain tumor (left) and liver tumor (right). For brain, a different MRI sequence (input image channel) is shown for each example.}
    \label{fig:translation}
\end{figure*}

Image translation and segmentation examples are shown in Fig.~\ref{fig:translation}, left. As shown in the figure, tumors were well removed by image-to-image translation. Furthermore, the residual output is similar to the segmentation output which supports our conjecture that this translation is similar to segmentation. Some of the sequences (T1, T1c, T2) result in fairly complicated residuals that are nonetheless correctly reinterpreted as segmentations via the residual decoder. The first column in Fig.~\ref{fig:translation} reveals an artifact of distribution imbalance where a rare truncated input slice is transformed into a common non-truncated slice. Artifacts of this sort are particularly common when there is an imbalance in the distribution of slice sizes between P and A (which we try to avoid). Ideally, entire brain volumes would be used as inputs instead of slices as performed here.

\subsection{Liver tumor segmentation}

Finally, a 2D liver tumor segmentation dataset was created with sets of healthy and diseased cases (with and without tumor, respectively) from the Liver Tumor Segmentation Challenge (LiTS) data~\cite{bilic2019liver}. The LiTS dataset contains portal venous phase contrast enhanced abdominal computed tomography (CT) scans along with pixel-level annotation of the liver and liver tumor boundaries. Inclusion criteria were the presence of colorectal cancer metastases; however, other tumors may be present but are not identified. Every CT volume corresponds to a single patient. The original data contains 130 CT volumes in the training set and 70 CT volumes in the heldout test set. To create our 2D dataset, we extracted and postprocessed relevant liver slices from only the publicly available training cases. The LiTS dataset is highly heterogenous, composed of cases collected from seven sites around the world, presenting vastly different anatomical features across cases; top methods submitted to the LiTS challenges predict tumor segmentation with inconsistent performance across different volumes~\cite{bilic2019liver,vorontsov2019deep}. Consequently, we performed a 4-fold cross-validation on our data to reduce the effect of this variability. Approximately 10\% of the data in every fold was used as a validation subset for hyperparameter selection and early stopping. Across training, validation, and testing subsets, no volumes were represented in more than one subset; models must generalize across cases.

Volumes were pre-processed by mean-centering the liver region in each volume and dividing it by five times its standard deviation, with statistics computed only in the liver. For each volume in the training set, we extracted axial slices with the constraint that the liver must cover at least 6\% of the slice area. Slices that contain no tumor are labeled as healthy; those with at least one tumor that is at least 6 pixels wide (along the coronal or sagittal directions) are labeled as diseased and the rest were discarded. This produced 6,319 healthy slices across 123 volumes and 11,165 diseased slices across 112 volumes. To focus on liver tumor segmentation and reduce the variability in liver size and location, slices were cropped to the liver and resized to 256$\times$256 pixels. The space outside of the liver is zero-filled and liver boundary annotations were discarded.

Because the liver dataset is composed of slices extracted from diseased volumes, the A (healthy) and P (diseased) domains differ not only in the absence vs presence of tumors but also in the distribution of axial positions from which slices are extracted. To reduce this bias, we sampled healthy slices during training according to a 20-bin histogram of the axial positions of diseased slices. In each epoch, the number of healthy slices sampled with replacement was equal to the total number of diseased slices sampled without replacement.

We trained the proposed model and baselines with reference annotations available for 1\% of the training data. As shown in Table~\ref{tab:results}, the proposed model achieves a $0.74$ Dice score, significantly outperforming the segmentation baseline ($0.66$) and the semi-supervised autoencoding ($0.66$) and mean teacher ($0.69$) baselines. Figure~\ref{fig:relative} shows how much of the gap is covered by each method between segmentation using 1\% of the data vs using 100\% of the data; viewed in this way, the proposed method covers 58\% $\pm$ 2\% of the gap, as compared to 22\% $\pm$ 7\% or 0\% $\pm$ 0\% by the mean teacher and autoencoding baselines, respectively. The autoencoding method shows a lack of improvement over fully supervised segmentation. The overall lower Dice scores for liver as compared to brain may be due to fewer training cases and less information in CT than in MRI scans. Image translation and segmentation examples are shown in Fig.~\ref{fig:translation}, right, where the unsupervised removal of tumors is evident.

\subsection{Ablations}

\begin{figure}[htb!]
    \centering
    \subcaptionbox{Brain}{
        \includegraphics[width=0.48\linewidth]{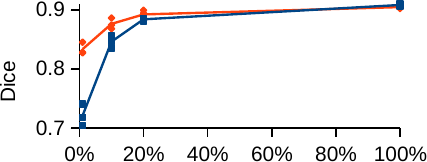}}
    \subcaptionbox{Liver}{
        \includegraphics[width=0.48\linewidth]{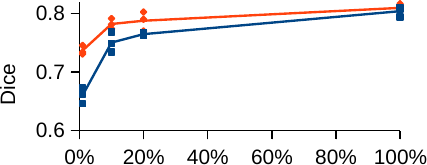}}
    \caption{Dice score when using reference annotations for various fractions (\%) of the data during training for (a) brain tumor dataset and (b) liver tumor dataset. Points: individual runs; lines: average across three runs. Orange, diamonds (top): proposed; blue, squares (bottom) supervised segmentation.}
    \label{fig:fractions}
\end{figure}

\textbf{Annotation efficiency.}
We tested the proposed method with reference tumor annotations provided for increasing fractions of the brain and liver training sets, as compared with the fully supervised segmentation baseline (Fig.~\ref{fig:fractions}). This demonstrates that the proposed method performs robustly with few labeled examples (1\%) and scales well with the proportion of unannotated data.

\textbf{Normalization.}
In Table~\ref{tab:ablations_normalization}, we compare the choice of different normalization layers in the encoder and the decoder of the proposed method, evaluated on the brain data. Using instance normalization in the encoder and layer normalization in the decoder (as proposed) or instance normalization in both yielded the best results.

\textbf{Absence to presence translation.}
As in prior work in image-to-image translation~\cite{zhu2017unpaired,liu2017unsupervised}, we perform translation both from the presence domain to the absence domain (P $\rightarrow$ A) and \textit{vice versa} (A $\rightarrow$ P). We conjecture that training P $\rightarrow$ A is sufficient; however, additionally training A $\rightarrow$ P makes more efficient use of the available data and should yield a better performing model. We confirm this in Table~\ref{tab:ablations_direction}. We note that when translation objectives are removed, performance(0.75 $\pm$ 0.02 Dice) on brain data is on par with the autoencoding and mean teacher baselines (probably because there are still reconstruction objectives).

\begin{table}[htb!]
\caption{Ablation experiments on (a) the normalization layers used in the encoder and the decoder of the proposed model (``IN'': instance norm, ``LN'': layer norm, ``BN'' batch norm~\cite{ioffe2015batch}; $\langle$ encoder : decoder $\rangle$) and (b) the directions of image-to-image translation, whether bidirectional between the presence (P) and absence (A) domains, or unidirectional only from P to A. Reporting tumour segmentation Dice score in brain: mean (standard deviation).}
\centering
\setlength{\tabcolsep}{6pt}
\begin{subtable}[t]{0.22\textwidth}
\begin{tabular}[t]{lc}
  Normalization & Dice \\
  \midrule
  IN : LN & \textbf{0.83 (0.01)} \\
  IN : IN & \textbf{0.83 (0.01)} \\
  LN : LN & 0.78 (0.02) \\
  BN : BN & 0.80 (0.03) \\ 
\end{tabular}
\caption{}
\label{tab:ablations_normalization}
\end{subtable}
\quad
\begin{subtable}[t]{0.22\textwidth}
\begin{tabular}[t]{lc}
  Translation & Dice \\
  \midrule
  P $\rightarrow$ A, A $\rightarrow$ P  & \textbf{0.83 (0.01)} \\ 
  P $\rightarrow$ A & 0.79 (0.02) \\
  no translation & 0.75 (0.02) \\
  \\
\end{tabular}
\caption{}
\label{tab:ablations_direction}
\end{subtable}
\end{table}

\textbf{Batch size.}
We found that the proposed method is not sensitive to batch size, after testing batch sizes of 5, 10, 15, 20, 25, and 30 on the validation subset of the brain tumor dataset.

\begin{figure}
    \centering
    \subcaptionbox{Brain}{
        \includegraphics[width=0.28\linewidth]{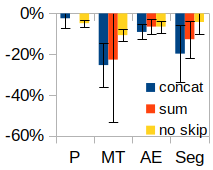}
    }
    \subcaptionbox{Liver}{
    \includegraphics[width=0.28\linewidth]{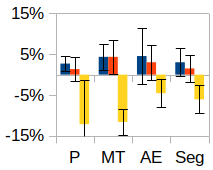}
    }
    \subcaptionbox{Synthetic hard}{
        \includegraphics[width=0.28\linewidth]{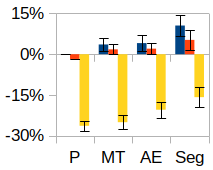}
    }
    \caption{The relative performance of the proposed model (``P''), mean teacher baseline (``MT''), autoencoding baseline (``AE''), and supervised segmentation baseline (``Seg'') for different long skip connection types (``concat'', blue; ``sum'', orange; ``no skip'', yellow), as compared to the proposed compressed skip connection. Using annotations for 1\% of the data.}
    \label{fig:long_skip_ablation}
\end{figure}

\textbf{Compressed skip connection.}
We evaluated the compressed skip connections proposed in Section~\ref{sec:long_skip}. In Figure~\ref{fig:long_skip_ablation}, we evaluate the performance of the proposed segmentation method, as well as every baseline method, on every task for different types of long skip connections: concatenation (``Concat''), summation (``Sum''), and no skips (``No skip''). Performance is evaluated relative the performance achieved with the proposed compressed skip connections. All models without skip connections perform poorly. The proposed compressed skip connections allow all baseline methods that we compare against to perform better on the brain data than do summation or concatenation type skip connections. For this data, when providing annotations for only 1\% of the volumes, the conventional skip connections produce highly variable results across training runs. For liver and synthetic data, models with compressed skip connections perform similarly to those with ‘sum’ and ‘concat’ skip connections for all models (proposed and baselines), with results typically falling within the standard deviation. We thus conclude that the proposed skip connections are the most consistent at producing quality results with low variance.

\textbf{Dual-use residual decoder.}
Finally, we compare in Table~\ref{tab:results} the use of a shared translation/segmentation decoder (``Proposed'') to models with a separate segmentation decoder (``sep dec'') for every task. The separate decoder performs on par with the shared decoder on the simple and large variants of the synthetic task and may slightly outperform the latter on the hard variant. However, a separate decoder underperforms on the brain and liver tumour segmentation tasks. Thus sharing decoders for segmentation and residual translation can be very beneficial (brain and liver data) but may not always be optimal.

% %%%%%%%%% Conclusion
\section{Extensions and applications}

Although we present work on two domains, P and A, the proposed method can be easily extended to any greater number of domains. For example, if different types of pathology are known to be present in a medical image dataset, a domain-specific code (with a corresponding residual decoder) could be encoded for each pathology in addition to a neutral code with all pathologies absent. Most interestingly, our image-to-image translation approach would allow any number of pathologies to be present in an image at a time, unlike for example the StarGAN multi-domain image-to-image translation architecture \cite{choi2018stargan}. This can be particularly useful for high-resolution histopathology imaging or x-ray multi-pathology segmentation.

Finally, we note that there are data outside of image segmentation that can be split into P and A domains. For example, in image registration problems where the dense deformation fields can be expressed based on the presence of tumors or morphological changes in the target images.

%material fault analysis, such as rust detection, microchip defects, or the decay of building facades can be expressed in this way. Another interesting application may be the surveying of flood damage by learning the difference between pre-flood and post-flood urban aerial images.

\section{Conclusion}
\label{sec:conclusion}

We propose a semi-supervised segmentation method that makes use of image-to-image translation in order to leverage unsegmented training data with cases presenting the object (P) of interest and cases in which it is absent (A). We argue that this objective is similar to segmentation because they both require disentangling the target object from the background. Indeed, we validate our method on brain tumor segmentation in MR images, liver tumor segmentation in CT images, as well as synthetic segmentation tasks, where we achieve significant improvements over supervised  and  semi-supervised baselines.

\section{Acknowledgements}

We would like to thank Ming-Yu Liu, Wonmin Byeon, Shalini De Mello, Varun Jampani, and Yoshua Bengio for taking the time to discuss this work. The compute resources from Nvidia and Compute Canada were essential for this work. We would also like to thank Francisco Perdigón Romero for providing reference code for mean teacher segmentation. 
% \input{acknowledgment}

%%%%%%%%%%%%%%%%%%%%%%%%%%
% Supplimentary Material %
%%%%%%%%%%%%%%%%%%%%%%%%%%
% \FloatBarrier
% \input{supplementary_materials}

%%%%%%%%%%%%%%%%%%%%%%%%%%
% Works cited            %
%%%%%%%%%%%%%%%%%%%%%%%%%%
\FloatBarrier
{\small
\bibliographystyle{IEEEtran}
\bibliography{biblio}
}

\end{document}